\documentclass[runningheads]{llncs}

\usepackage{eccv}

\usepackage{eccvabbrv}

\usepackage{graphicx}
\usepackage{booktabs}
\usepackage{adjustbox}
\usepackage{multirow}
\usepackage{float}

\usepackage[accsupp]{axessibility}  

\usepackage{multirow}
\usepackage{array}
\usepackage{pifont}
\newcommand{\cmark}{\ding{51}}

\makeatletter
\newcommand{\equalcontrib}[1]{%
  \textsuperscript{\@fnsymbol{#1}}%
}
\makeatother

\usepackage[pagebackref,breaklinks,colorlinks,citecolor=eccvblue]{hyperref}
\usepackage{hyperref}
\hypersetup{
    colorlinks=true,
    linkcolor=blue,
    filecolor=magenta,      
    urlcolor=blue,
    pdftitle={OmniDS Project Page},
}

\usepackage{orcidlink}

\begin{document}

\title{OmniDS: Dual-Stream Context Fusion for Omnidirectional Depth from Fisheye Cameras} 

\titlerunning{OmniDS}

\author{Chaesong Park\inst{1}\thanks{Equal contribution.}\orcidlink{0009-0007-1815-236X} \and
Jihyeon Hwang\inst{1}\equalcontrib{1}\orcidlink{0009-0008-0703-7953} \and
Muyeol Sung\inst{1}\orcidlink{0009-0000-2676-383X} \and
Jongwoo Lim\inst{1}\thanks{Corresponding author.}\orcidlink{0000-0002-2814-4765}}

\authorrunning{C.~Park et al.}

\institute{Seoul National University, Seoul, Republic of Korea\\
\email{\{chase121, zhyeon, smy4024169, jongwoo.lim\}@snu.ac.kr}}

\maketitle

\begin{abstract}
Omnidirectional depth estimation from multi-fisheye camera rigs is complicated by visibility conflicts: wide baselines cause different cameras to observe different portions, or even different faces, of the same object, so aggregating their features into a unified equirectangular (ERP) representation under fixed projection produces ambiguous matching evidence near occlusion boundaries and thin structures. Although existing methods mitigate this by down-weighting unreliable views, they do not resolve the underlying discrepancy because context formation and cross-view fusion remain tied to rigid fisheye-to-ERP sampling. 
We present \textbf{OmniDS}, an iterative depth refinement framework that replaces rigid aggregation by combining dynamic context fusion with consensus-aware multi-view similarity. A dual-stream encoder pairs a lightweight CNN for geometric detail with a frozen DINOv3 for semantic priors; their features are reprojected into ERP space at each refinement step via learned view weighting and deformable cross-attention with geometric distortion bias. In parallel, a multi-view consensus volume captures global cross-camera agreement through group-wise correlation and feature variance, regularized by a 3D U-Net. For efficient deployment, we distill the dual-stream representation into a single MobileNet-based encoder. OmniDS achieves state-of-the-art performance on the OmniThings, OmniHouse, and Sunny benchmarks while maintaining competitive inference speed. Project page and codes are available \href{https://parkchaesong.github.io/omnids}{here}.

  \keywords{Omnidirectional depth estimation \and Multi-view stereo \and Iterative refinement \and Cross-modal feature}
\end{abstract}

\section{Introduction}
\label{sec:intro}

\begin{figure}[t]
    \centering
    \includegraphics[width=1.0\linewidth]{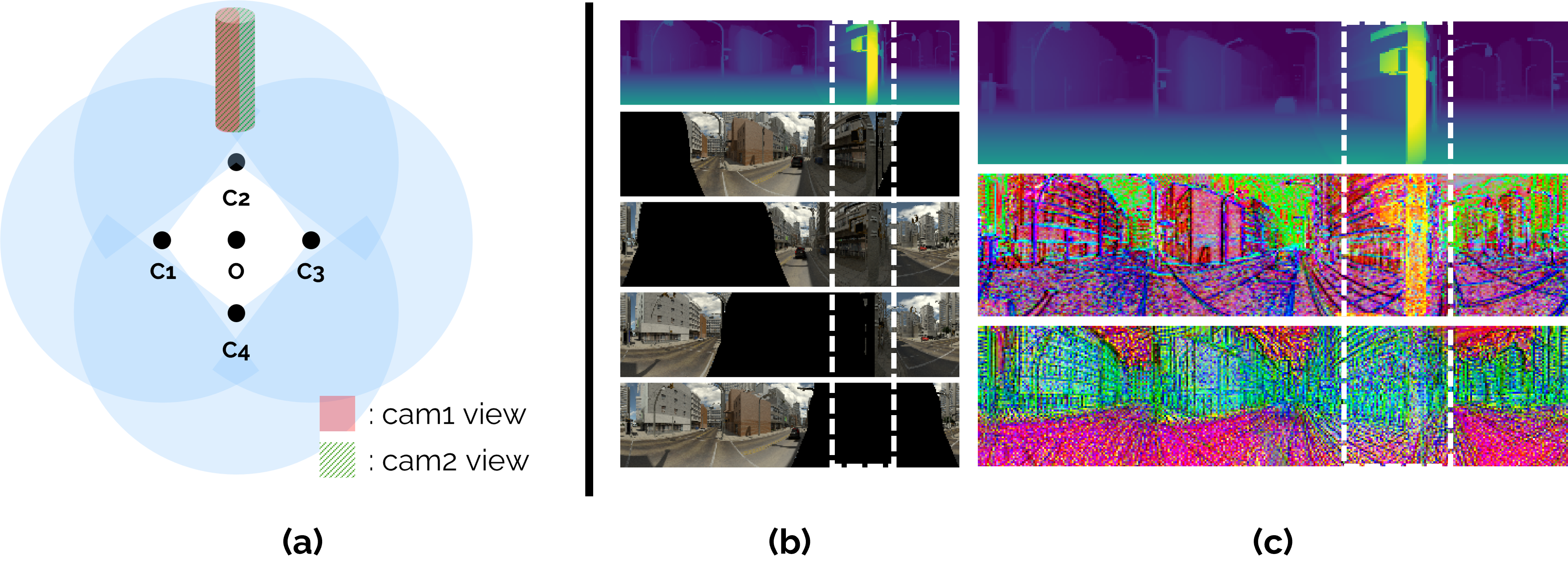}
    \caption{
    \textbf{(a)} A wide-baseline multi-fisheye rig induces strong parallax and self-occlusion: different cameras observe different visible faces of the same nearby object.
    \textbf{(b)} \textbf{Top:} ground-truth depth. \textbf{Below:} four per-camera renderings obtained by projecting each fisheye image into a common rig-centric ERP coordinate system using the GT depth. Even with perfect geometry, ERP-aligned samples can be inconsistent due to visibility conflicts (occlusion leakage) and self-occlusion (different surface faces mapped to the same ERP cell).
    \textbf{(c)} \textbf{Top:} ground-truth depth. \textbf{Middle:} the ERP context produced by our dynamic ERP context fusion. \textbf{Bottom:} a baseline ERP context built by simple grid sampling after fisheye-to-ERP projection. Our fusion yields a more coherent context around thin structures and occlusion boundaries.}
    \label{fig:intro}
\end{figure}

Omnidirectional depth perception is a cornerstone of reliable robotic autonomy.
Dense depth around the platform enables safe obstacle avoidance, local motion planning, and interaction in close-range clutter, and it also underpins long-horizon mapping and localization (e.g., SLAM) in indoor service robots, warehouse/industrial inspection, and AR/VR telepresence systems \cite{9812203}.
In these settings, blind spots and narrow fields-of-view are not merely inconvenient \cite{de2018eliminating}: they can lead to missed obstacles, unstable tracking, or incorrect trajectory decisions when the robot navigates tight corridors, crowded spaces, or dynamic human environments. Therefore, perceiving 360$^\circ$ depth with minimal occlusions and seamless coverage has become increasingly important for real-world deployment.

This work focuses on omnidirectional depth estimation with multi-fisheye camera rigs, a practical sensing configuration that provides dense measurements with full-surround coverage in a compact and low-cost form factor.
Such surround-view sensing is increasingly common in large-scale datasets and benchmarks, \cite{liao2022kitti, yogamani2019woodscape, zayene2025helvipad}, further motivating robust learning-based omnidirectional geometry perception.

However, learning-based omnidirectional depth from multi-fisheye inputs remains challenging due to severe lens distortion, heterogeneous overlaps among cameras, and wide-baseline parallax that breaks simple correspondence assumptions after projection to a unified rig-centric Equirectangular Projection (ERP).
To address these challenges, existing learning-based solutions broadly follow three paradigms:
(1) \textbf{Volumetric methods} (e.g., OmniMVS~\cite{won2019omnimvs}, CasOmniMVS~\cite{CasOmniMVS}) that sweep/warp multi-view features onto concentric spheres, but often incur heavy 3D regularization and memory cost;
(2) \textbf{Iterative refinement} (e.g., RomniStereo~\cite{jiang2024RomniStereo}) that adapts RAFT-style~\cite{teed2020raft} recurrent updates to omnidirectional geometry, yet typically relies on rigid cross-view aggregation under fixed projection rules; and
(3) \textbf{Rectification-based schemes} (e.g., OmniVidar~\cite{xie2023omnividar}, OmniStereo~\cite{deng2025omnistereo}, MODE~\cite{MODE}) that decompose omnidirectional estimation into multiple tractable stereo sub-problems.

While these paradigms differ in how they construct matching evidence, most of them rely on rigid, geometry-driven warping/rectification followed by local similarity after projection.
However, wide-baseline rigs induce severe parallax and self-occlusion, so different cameras often observe different visible portions (or even different faces) of the same object (\cref{fig:intro}a).
Consequently, aggregating evidence in a rig-centric ERP can be inherently multi-modal near thin structures and occlusion boundaries: even with accurate depth for projection, ERP-aligned samples can conflict due to (i) occlusion leakage, where an occluded camera contributes background pixels, and (ii) self-occlusion, where visible cameras map distinct surface faces (with different appearance and normals) into the same ERP cell (\cref{fig:intro}b).
This feature discrepancy creates ambiguous or contradictory matching signals; pixel-wise similarity and small local context tend to average incompatible evidence, yielding unstable updates exactly where visibility discontinuities dominate.

Prior works partially mitigate this issue by reducing the influence of unreliable views, e.g., using view-dependent weighting (RomniStereo~\cite{jiang2024RomniStereo}) or center-view-based processing (MDP-Omni~\cite{son2025mdp}).
While effective in suppressing gross outliers, such strategies primarily down-weight conflicting measurements; they do not explicitly resolve the underlying multi-modal discrepancy that arises when distinct surface characteristics are aggregated under a fixed fisheye-to-ERP projection.
As a result, iterative refinement remains bottlenecked by two rigid components: (i) context formation that is weak in textureless regions when built from CNN features alone, and (ii) fixed projection/aggregation that cannot adapt sampling and fusion across visibility discontinuities and thin-structure boundaries.

We present \textbf{OmniDS}, an iterative omnidirectional depth framework that avoids rigid ERP-aligned aggregation by combining dynamic ERP context fusion with consensus-aware multi-view similarity for robust refinement.

Our key contributions are summarized as follows:
\begin{itemize}
\item \textbf{Dynamic ERP Context Fusion for Fisheye Stereo.}
We propose a dual-stream representation that combines a CNN encoder for geometric matching with a frozen DINOv3 \cite{simeoni2025dinov3} backbone for semantic priors. This design enables robust context construction under fisheye distortions and occlusions.

\item \textbf{Multi-View Correlation and Consensus Volumes.}
We introduce a similarity representation tailored for omnidirectional multi-view stereo. Beyond a spatially aggregated correlation profile, we build a multi-view consensus volume that fuses group-wise correlation with cross-camera feature variance to enforce global agreement across cameras. Regularized by a 3D U-Net in a multi-scale pyramid, it provides stable geometric cues for iterative depth refinement.

\item \textbf{Efficient Deployment via Distilled OmniDS.}
To reduce inference cost, we distill the dual-stream representation into a single MobileNet-based encoder while preserving the geometric and semantic characteristics learned during training. The distilled model maintains near state-of-the-art accuracy while enabling real-time inference at approximately 10 FPS on a RTX 4070 SUPER.

\item \textbf{State-of-the-Art Performance.}
Our method achieves state-of-the-art performance across the OmniThings, OmniHouse, and Sunny benchmarks, outperforming prior methods on the majority of evaluation metrics. Comprehensive ablation studies further validate the contribution of each proposed component.

\end{itemize}

\section{Related works}

\subsection{Omnidirectional depth estimation}
Omnidirectional depth estimation has progressed through the development of correspondence formulations defined in spherical coordinate systems. A canonical formulation underlying many approaches is \textbf{depth-hypothesis alignment}: features from multiple cameras are warped under discrete depth hypotheses to produce depth-indexed matching evidence.

Within this paradigm, early learning-based methods explicitly adopted this paradigm by constructing hypothesis-aligned volumes and predicting depth from aggregated costs~\cite{won2019sweepnet,won2019omnimvs}, while later work primarily improved efficiency or the regularization/refinement scheme without abandoning depth-conditioned evidence. Meuleman et al.~\cite{meuleman2021real} preserved the sweeping-based correspondence formulation while re-engineering the cost computation and filtering stages to enable real-time execution, emphasizing efficiency in depth-wise evidence construction.

Later studies have pursued improved efficiency and robustness while largely retaining the central notion of forming depth-conditioned matching evidence. Xie et al.~\cite{xie2023omnividar} decomposed omnidirectional matching into rectified stereo sub-problems and fused the results. Jiang et al.~\cite{jiang2024RomniStereo} replaced heavy 3D cost regularization with iterative 2D recurrent refinement, and Deng et al.~\cite{deng2025omnistereo} further pursued real-time inference via simplified projection and confidence-guided fusion. Son et al.~\cite{son2025mdp} addressed oversmoothing from unimodal depth assumptions by using multimodal depth-prior sampling to adapt the search range across stages.

Despite these advances, most approaches still rely mainly on local matching cues from depth-swept feature sampling, which can be brittle around thin structures, object boundaries, and textureless regions, especially under severe fisheye distortions. We address this limitation by incorporating a global context representation that complements local matching evidence during depth refinement.

\subsection{Dual-stream feature representations for robust matching}
Recent work on dense correspondence suggests that local detail and global semantics are complementary. Frequency analyses help explain why: ViT self-attention tends to act as content-adaptive smoothing that suppresses high-frequency signals, while convolutions emphasize high-frequency components~\cite{park2022how,bai2022revisitinghf}. Beyond frequency, Naseer et al.~\cite{naseer2021intriguing} show that ViTs are strongly robust to severe occlusions and significantly less biased toward local textures than CNNs, indicating that transformer representations can emphasize global shape and context. In parallel, self-supervised vision foundation models like DINO expose object layout and boundaries directly in the last-layer attentions~\cite{caron2021dino}, serving as a strong source of semantic priors.

These observations have recently been operationalized in dense matching \cite{sun2021loftr, wang2022matchformer}. RoMa~\cite{edstedt2024roma} leverages frozen DINO features for robust coarse matching, but explicitly compensates for their limited spatial precision by combining them with specialized ConvNet fine features to form a localizable feature pyramid. A Tale of Two Features~\cite{zhang2023tale} similarly demonstrates that fusing foundation-model representations (DINOv2) with generative-model features improves zero-shot semantic correspondence, reinforcing that different pretrained representations contribute different strengths. Related trends also appear in stereo depth estimation: FoundationStereo~\cite{wen2025foundationstereo} adapts internet-scale monocular priors from a foundation model into stereo to improve zero-shot generalization.

Motivated by these observations, recent work increasingly adopts dual-stream representations that combine convolutional features for precise localization with transformer features for semantic context. In line with this, our work utilizes DINOv3 features not just as simple semantic descriptors, but as a robust prior to navigate the severe distortions inherent in omnidirectional systems. Such representations are particularly beneficial for omnidirectional imagery, where equirectangular projection introduces significant distortions \cite{jung2025edm} and increases the need for both semantic context and precise local geometry.

\subsection{Geometry encoding volume and iterative refinement}
Cost volume construction and iterative refinement have become the dominant paradigm in modern stereo matching. Early approaches such as GwcNet \cite{guo2019group} showed that the structure of the cost volume critically affects matching accuracy. Instead of collapsing features into a single correlation channel, group-wise correlation divides channels into groups and computes per-group similarities, preserving richer discriminative information while maintaining computational efficiency. The resulting volume is regularized with a 3D stacked hourglass network to aggregate spatial context.

Subsequent works explored more expressive cost representations. ACVNet \cite{xu2022attention} introduced attention-weighted concatenation volumes, where correlation-derived attention selectively filters concatenated features to produce more informative and compact cost encodings. These methods highlight that cost volume design is not merely a similarity computation, but a structured geometric representation.
Iterative refinement has further reshaped stereo matching. RAFT-Stereo~\cite{lipson2021raft} adapts RAFT~\cite{teed2020raft} by building an all-pairs correlation volume and refining disparity via recurrent updates with correlation lookup, showing that iterative optimization can replace heavy 3D cost regularization. IGEV-Stereo~\cite{xu2023iterative} extends this idea with a Geometry Encoding Volume regularized by a lightweight 3D hourglass, which initializes disparity and guides each refinement step.

However, these methods target rectified binocular stereo with a 1D disparity search. In omnidirectional multi-fisheye settings, (i) the search is over inverse depth along spherical rays, (ii) evidence must be jointly encoded from multiple cameras with different baselines and overlaps, and (iii) fisheye distortion yields spatially varying feature reliability that standard cost volumes do not explicitly model.

\begin{figure}[h]
    \centering
    \includegraphics[width=1.0\linewidth]{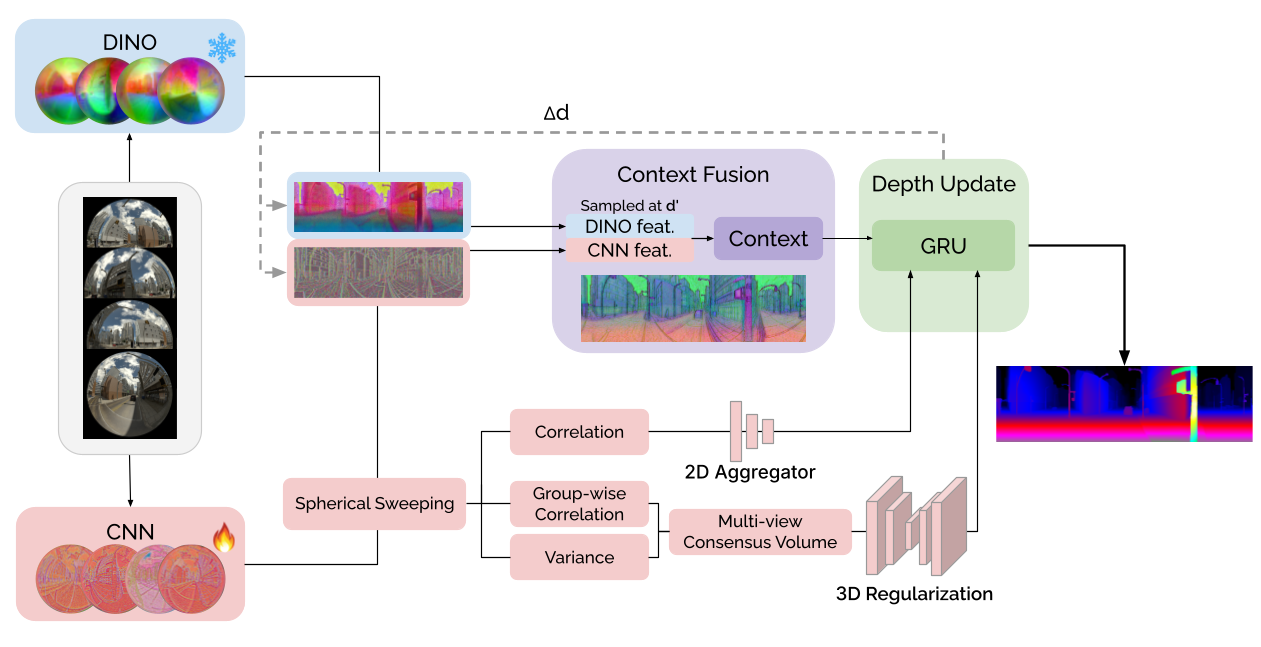}

\caption{\textbf{Overview of OmniDS framework.} Our framework takes four fisheye images and iteratively refines an ERP inverse depth map. The current depth estimate is denoted as $d'$, and the ConvGRU predicts a residual update $\Delta d$ to refine it. The model leverages a dual-stream encoder to extract semantic and geometric features, constructing both dynamic context and static similarity volumes for robust depth estimation.} 
   \label{fig:overall}
\end{figure}
\section{Method}  
\subsection{Overview}
Given four multi-view fisheye images ${I_k}_{k=1}^{4}$ with known camera models, we estimate an ERP inverse depth map $\hat{d}\in\mathbb{R}^{H\times W}$ (Fig.~\ref{fig:overall}). OmniDS first extracts per-view features with a dual-stream encoder: a lightweight CNN for geometric matching and a frozen DINOv3 ViT for semantic context. The CNN and DINO features are then projected to ERP space (via grid sampling with learned view weights and deformable cross-attention with distortion bias, respectively) and fused pointwise to form the context representation.

In parallel, the CNN features are swept into spherical volumes to construct both a correlation profile as a local matching signal with spatial aggregation and a multi-view consensus volume as a globally regularized depth cost via group-wise correlation and multi-view variance, processed by a 3D regularization network. A ConvGRU update module then iteratively refines the inverse depth estimate, receiving the fused context, correlation profile, and consensus volume features at each iteration.

\subsection{Feature Extraction}
Each fisheye image $I_k$ is processed by a dual-stream architecture to extract complementary representations. A lightweight CNN encoder shared across cameras produces feature maps $F_k^{\text{cnn}}$ that capture fine-grained textures and local edges, which are essential for high-frequency stereo matching and the construction of the multi-view consensus volume. In parallel, a frozen DINOv3 ViT backbone followed by a learnable projection head produces $F_k^{\text{dino}}$ that encode semantic and structural priors, enabling the network to resolve depth ambiguities in challenging regions such as textureless surfaces, reflections, or complex occlusions where local matching often fails. The CNN stream is used for all geometric operations-correlation, group-wise correlation, and ERP context-while the DINO stream is reserved for semantic context via cross-attention.


While the dual-stream approach provides superior accuracy during training, running both a CNN and a ViT at inference time is computationally expensive. To bridge this gap, we adopt a knowledge distillation strategy during fine-tuning stage. We distill the geometric characteristics of the CNN stream and the semantic richness of the DINO stream into a single, unified MobileNet-based backbone. This distilled encoder serves as the final feature extractor during inference, significantly reducing latency and memory footprint while preserving the performance gains of the original dual-stream design. The details of this distillation procedure are provided in \cref{Distillation}.

\subsection{Context Fusion} \label{contextfusion}

Unlike static volumes, the context representations are dynamically updated at each iteration $t$ to focus on the most relevant features given the current inverse depth estimate $\hat{d}^{(t)}$. This module integrates semantic priors from the DINO stream and geometric details from the CNN stream into a unified ERP-space representation.

For each ERP pixel $i$, we project its 3D position-determined by the current depth $\hat{d}_i^{(t)}$-onto the four fisheye views. We sample the corresponding CNN features $F_{i,k}^{\mathrm{cnn}}$ using grid sampling. To robustly fuse these multi-view features, we employ a learned weighting mechanism. 
We concatenate the CNN features from all cameras and pass them through a lightweight network $\textit{cam}_{\textit{weight}}$ to predict camera-specific logits, which are then normalized via a Softmax layer to produce view weights $\mathbf{w}_{i,k}$:
\begin{equation}
w_{i,k} = 
\mathrm{Softmax}_k \Big(
\mathrm{Cam}_{\mathrm{weight}}
\big(
[F_{i,1}^{\mathrm{cnn}}; F_{i,2}^{\mathrm{cnn}}; F_{i,3}^{\mathrm{cnn}}; F_{i,4}^{\mathrm{cnn}}]
\big)
\Big).
\end{equation}

The final CNN-based ERP feature is computed as a weighted sum:
\begin{equation}
F_i^{\mathrm{cnn\text{-}erp}} = \sum_{k=1}^{4} w_{i,k} F_{i,k}^{\mathrm{cnn}} .
\end{equation}

To incorporate wider structural context, we project the DINO features $F_{k}^{\mathrm{dino}}$ using deformable cross-attention \cite{zhu2020deformable}. Each query $\mathbf{q}_i$ is a learnable ERP embedding augmented with a depth encoding of the current estimate $\hat{d}_i^{(t)}$. 
For each view $k$, we project the 3D point at $\hat{d}_i^{(t)}$ onto the fisheye image to obtain a reference point $\mathbf{p}_{i,k}$ for the ERP query $\mathbf{q}_i$. 
Instead of simple grid sampling, the ERP query attends to $P$ learned sampling offsets $\Delta \mathbf{p}_{i,k,j}$ around the reference point:
\begin{equation}
\mathrm{CrossAttn}(\mathbf{q}_i) =
\sum_{k=1}^{4} \sum_{j=1}^{P}
A_{i,k,j} \cdot
\mathbf{W}_v \,
F_k^{\mathrm{dino}}
\big(
\mathbf{p}_{i,k} + \Delta \mathbf{p}_{i,k,j}
\big),
\label{eq:crossattn}
\end{equation}

where $A_{i,k,j}$ are attention weights. 
\begin{equation}
b(r_{i,k}) = \mathrm{MLP}\left(
\min\left(
\left\| 2\mathbf{p}_{i,k} - \mathbf{1} \right\|_2,\,
1
\right)
\right).
\label{eq:bias}
\end{equation}
To account for the non-uniform resolution of fisheye optics, we add a geometric distortion bias $b(r_{i,k})$ to the attention logits(Eq.~\eqref{eq:bias}), where $r_{i,k}$ is the radial distance from the camera center. This encourages the model to rely more on the high-resolution central regions of the fisheye images. The aggregated outputs $\mathrm{CrossAttn}(\mathbf q_i)$ over all ERP queries form the DINO-projected ERP feature map $F^{\mathrm{dino\text{-}erp}}$.

To obtain the final context $F^{\mathrm{context}}$ , we integrate the ERP-aligned DINO features $F^{\mathrm{dino\text{-}erp}}$ and the CNN-based ERP features $F^{\mathrm{cnn\text{-}erp}}$ using a residual pointwise convolution:
\begin{equation}
F^{\mathrm{context}} =
F^{\mathrm{dino\text{-}erp}}
+
\mathrm{Fuse}
\big(
[
F^{\mathrm{dino\text{-}erp}};
F^{\mathrm{cnn\text{-}erp}}
]
\big).
\end{equation}

This design ensures that semantic information from DINO is preserved as a strong prior, while CNN features provide the necessary geometric refinement.

\subsection{Similarity Volumes}
\label{similarityvolumes}

While the fused context provides a dynamic representation of the scene, we pre-compute static similarity volumes across $D$ discrete depth bins to provide stable geometric constraints. These volumes act as a global cost signal that guides the iterative refinement process.

\paragraph{Correlation profile.}

Following the multi-view arrangement in RomniStereo \cite{jiang2024RomniStereo}, we organize the CNN spherical sweep volumes $\{V_k\}_{k=1}^{4}$ into reference and target pairs. For each pair, a correlation profile $S(h,w,d)$ is computed via channel-wise correlation:
\begin{equation}
S(h,w,d) =
\frac{1}{\sqrt{C}}
\left\langle
V^{\mathrm{ref}}_{:,h,w,d},
V^{\mathrm{tgt}}_{:,h,w,d}
\right\rangle .
\end{equation}

To ensure that the matching signal is spatially consistent, we apply a 2D spatial aggregator consisting of several convolution layers. This allows the similarity at a specific depth to be influenced by its neighbors, smoothing local noise and reducing ambiguities in repetitive textures.

\paragraph{Multi-View Consensus Volume.}

To fully exploit the overlapping fields of view from all four cameras, we construct a consensus volume that captures global agreement. This volume consists of two primary signals. 

First, we compute group-wise correlations by dividing the feature channels into groups and computing correlations within each group, as in \cite{guo2019group, xu2022attention}. This preserves a multi-modal distribution of matching costs and provides a richer representation than a single scalar correlation.

Second, we compute the variance of features across all four cameras at each ERP-depth coordinate. This signal provides a strong indicator of depth accuracy: the variance becomes small when all cameras agree on the feature appearance (i.e., the correct depth) and increases near occlusion boundaries or incorrect depth hypotheses.
These two signals are concatenated to form a raw consensus volume. Unlike the correlation volume, this volume is regularized using a 3D U-Net. By performing convolutions across both the spatial $(H,W)$ and depth $(D)$ dimensions, the network enforces global geometric consistency and helps the model escape local minima in the cost space.

Both the correlation volume and the multi-view consensus volume are organized into a 4-level multi-scale pyramid, allowing the ConvGRU to retrieve the relevant costs at the current depth $\hat{d}^{(t)}$ during each refinement step $t$.

\subsection{Iterative Depth Refinement}

The inverse depth is initialized at the midpoint of the depth range and iteratively refined using a ConvGRU. At each iteration $t$, the context is recomputed at the current estimate. Specifically, DINO features are reprojected via cross-attention, CNN features are sampled via grid sampling, and the two are fused to form the contextual representation. The correlation profile and consensus volume features are queried at the current depth.

A motion encoder aggregates these signals into a motion feature $\mathbf{m}^{(t)}$, and the GRU updates its hidden state:
\begin{equation}
\mathbf{h}^{(t)} =
\mathrm{GRU}
\big(
\mathbf{h}^{(t-1)},\,
[F^{\mathrm{context},(t)};\,\mathbf{m}^{(t)}]
\big).
\end{equation}

A depth head predicts a residual update $\Delta d^{(t)}$ from the hidden state and refines the estimate by adding it to the previous prediction, i.e., $\hat{d}^{(t)} = \hat{d}^{(t-1)} + \Delta d^{(t)}$. The final output $\hat{d}_{\mathrm{up}}^{(t)}$ is obtained via convex upsampling, which upsamples the prediction by $2\times$ along the spatial dimensions and the depth-bin axis.

We supervise all iterations using a weighted sequence loss following \cite{lipson2021raft}:
\begin{equation}
\mathcal{L}
=
\sum_{t=1}^{T}
\gamma^{T-t}
\left\|
\hat{d}_{\mathrm{up}}^{(t)} - d^{*}
\right\|_{1},
\end{equation}
where $d^{*}$ denotes the ground-truth depth and $\gamma = 0.9$ emphasizes later iterations.

\subsection{Feature Distillation} \label{Distillation}
To reduce feature-extraction cost, we distill two frozen teacher encoders (CNN features and DINO context features) into a single lightweight student encoder.
The student is a U-Net-style MobileNetV2 that outputs two feature maps via $1\times1$ heads: a $\times2$ map replacing CNN features and a $\times16$ map replacing DINO features, both with 32 channels for compatibility.

We train the student with a weighted feature-level MSE:
\begin{equation}
\mathcal{L}_{\text{distill}}=\lambda_{\text{cnn}}\mathrm{MSE}(\mathbf{F}^{s}_{\text{cnn}},\mathbf{F}^{t}_{\text{cnn}})
+\lambda_{\text{dino}}\mathrm{MSE}(\mathbf{F}^{s}_{\text{dino}},\mathbf{F}^{t}_{\text{dino}}),
\end{equation}
using $\lambda_{\text{cnn}}=1.0$ and $\lambda_{\text{dino}}=10.0$ to emphasize the harder-to-match DINO representations.
Distillation optimizes only the student with AdamW (lr $5\times10^{-4}$, wd $10^{-5}$) for 1 epoch on OmniThings.
Afterwards, the student replaces both teachers in the pipeline, and we fine-tune the full model end-to-end on OmniHouse and Sunny for 15 epochs.

\section{Experiments}

\subsection{Datasets and Evaluation}
\paragraph{Datasets.}
We evaluate on the synthetic omnidirectional benchmarks used in prior work, namely OmniThings, OmniHouse, Sunny, Cloudy, and Sunset~\cite{won2019omnimvs, won2019sweepnet}.
Each sample contains four $220^\circ$ fisheye images (front/right/back/left) at $768\times800$ and a GT inverse-depth map on the equirectangular sphere at $360\times640$ (with $\phi\!\in[-\pi/2,\pi/2]$, $\theta\!\in[-\pi,\pi]$).
OmniThings provides large-scale object-centric diversity, OmniHouse focuses on realistic indoor environments, and Sunny/ Cloudy/ Sunset are outdoor driving scenes sharing the same layouts with varying weather.

\paragraph{Training and evaluation protocol.}
We first train our model using OmniThings only.
After the initial training stage as in \cite{jiang2024RomniStereo, son2025mdp}, we perform distillation-based fine-tuning on a mixed dataset composed of OmniHouse and Sunny, following the procedure described in \cref{Distillation}.
For evaluation, we report results on the test sets of all five datasets to assess both in-domain performance and cross-domain robustness across indoor/outdoor scenes and varying weather conditions.

\paragraph{Evaluation metrics.}
Following prior omnidirectional stereo/depth works, we evaluate predictions in the inverse index space.
Specifically, we discretize the inverse depth into $N$ predefined indices and measure the percent error of the predicted inverse index with respect to the GT, normalized by $N$.
We report the percentage of pixels whose index error is larger than $1$, $3$, and $5$ (denoted as $>1$, $>3$, and $>5$), as well as the mean absolute error (MAE) and root mean square error (RMS) computed in the inverse index domain.

\subsection{Implementation Details}

\paragraph{Model Configuration.}
 The input consists of four $768 \times 800$ fisheye images. Our dual-stream encoder combines an 18-layer residual CNN with a frozen DINOv3. While the final output uses $N=192$ inverse depth hypotheses, the model predicts $96$ depth bins within the refinement loop, which are subsequently upsampled by a factor of two via convex upsampling along the depth axis. The iterative refinement begins with the initial inverse depth set to the midpoint of these bins (i.e., $48$). The deformable cross-attention module is configured with 4 heads and 4 sampling points each. For geometry encoding, we construct a correlation profile and a multi-view consensus volume using 8-group correlation, both utilizing 4-level pyramids with a lookup radius of 4 to produce 36-dimensional feature vectors.

\paragraph{Training Schedule.} The model is first pre-trained on the OmniThings dataset for 30 epochs. Subsequently, both the feature distillation and the final fine-tuning on the OmniHouse and Sunny datasets are conducted for 15 epochs each.

\subsection{Quantitative Results}

\begin{table*}[t]
\centering
\scriptsize
\setlength{\tabcolsep}{4pt}
\renewcommand{\arraystretch}{1.2}
\caption{Quantitative comparison on OmniThings and OmniHouse datasets.}
\begin{adjustbox}{max width=\textwidth}
\begin{tabular}{l|ccccc|ccccc|c}
\hline
\multirow{2}{*}{Method}
& \multicolumn{5}{c|}{OmniThings}
& \multicolumn{5}{c|}{OmniHouse}
& \multirow{2}{*}{\shortstack{Time\\(ms)}} \\
\cline{2-11}
& \multicolumn{1}{c}{>1} & \multicolumn{1}{c}{>3} & \multicolumn{1}{c}{>5} & \multicolumn{1}{c}{MAE} & \multicolumn{1}{c|}{RMS}
& \multicolumn{1}{c}{>1} & \multicolumn{1}{c}{>3} & \multicolumn{1}{c}{>5} & \multicolumn{1}{c}{MAE} & \multicolumn{1}{c|}{RMS} & \\
\hline

\multicolumn{12}{l}{\textit{Trained on OmniThings only}} \\
\hline
OmniMVS~\cite{won2019omnimvs}
& 47.72 & 15.12 & 8.91 & 2.40 & 5.27
& 30.53 & 10.29 & 6.27 & 1.72 & 4.05
& 289 \\
OmniMVS+$_{32}$~\cite{won2020end}
& 20.70 & 8.18 & 5.49 & 1.37 & 4.11
& 19.89 & 5.89 & 3.99 & 1.30 & 2.64
& 289 \\
S-OmniMVS~\cite{chen2023s}
& 28.03 & 10.40 & 6.33 & 1.48 & \underline{3.68}
& 18.86 & 8.05 & 4.90 & 1.06 & 2.41
& -- \\
RomniStereo$_{32}$~\cite{deng2025omnistereo}
& 20.42 & 8.49 & 5.81 & 1.39 & 4.22
& 12.13 & 4.73 & 3.02 & 0.80 & 1.85
& \textbf{83} \\
RomniStereo$_{64}$~\cite{deng2025omnistereo}
& \underline{17.77} & 7.52 & 5.00 & 1.22 & 3.90
& \underline{10.52} & \underline{4.05} & \underline{2.69} & \underline{0.74} & \underline{1.73}
& 161 \\
MDP-Omni~\cite{son2025mdp}
& 18.37 & \underline{7.09} & \underline{4.59} & \underline{1.20} & 3.79
& 17.13 & 7.20 & 4.68 & 1.16 & 2.62
& \underline{117} \\
Ours
& \textbf{14.89} & \textbf{5.94} & \textbf{3.96} & \textbf{1.04} & \textbf{3.59}
& \textbf{9.42} & \textbf{3.28} & \textbf{2.01} & \textbf{0.64} & \textbf{1.61}
& 148 \\
\hline

\multicolumn{12}{l}{\textit{Fine-tuned on OmniHouse and Sunny}} \\
\hline
OmniMVS-ft~\cite{won2019omnimvs}
& 50.28 & 22.78 & 15.60 & 3.52 & 7.44
& 21.09 & 4.63 & 2.58 & 1.04 & 1.97
& 289 \\
OmniMVS+$_{32}$-ft~\cite{won2020end}
& 44.79 & 27.17 & 20.41 & 4.23 & 8.42
& 9.70 & 3.51 & 2.13 & 0.64 & 1.69
& 289 \\
S-OmniMVS-ft~\cite{chen2023s}
& -- & -- & -- & -- & --
& 6.99 & 1.79 & 0.97 & \underline{0.42} & 1.06
& -- \\
RomniStereo$_{32}$-ft~\cite{deng2025omnistereo}
& 34.32 & 19.76 & 14.22 & 2.81 & 6.47
& 6.02 & 2.49 & 1.73 & 0.49 & 1.31
& \textbf{83} \\
RomniStereo$_{64}$-ft~\cite{deng2025omnistereo}
& 29.84 & 16.21 & 11.28 & 2.26 & 5.60
& 5.28 & 2.22 & 1.51 & \underline{0.42} & 1.14
& 161 \\
MDP-Omni-ft~\cite{son2025mdp}
& 29.53 & 14.05 & \underline{9.12} & \underline{1.96} & \underline{5.08}
& 5.61 & 2.19 & 1.50 & 0.44 & 1.15
& 117 \\
Ours-ft
& \textbf{25.14} & \textbf{11.72} & \textbf{7.67} & \textbf{1.70} & \textbf{4.69}
& \underline{3.92} & \underline{1.17} & \textbf{0.64} & \textbf{0.28} & \textbf{0.89}
& 148 \\
Ours-ft (distilled)
& \underline{26.22} & \underline{13.69} & 9.51 & 2.11 & 5.49
& \textbf{3.41} & \textbf{1.16} & \underline{0.73} & \textbf{0.28} & \textbf{0.89}
& \underline{102} \\
\hline
\end{tabular}
\end{adjustbox}
\label{tab:quantitative_omnithings_house}
\end{table*}

\begin{table*}[t]
\centering
\scriptsize
\setlength{\tabcolsep}{4pt}
\renewcommand{\arraystretch}{1.2}
\caption{Quantitative comparison on Sunny, Cloudy, and Sunset datasets.}
\begin{adjustbox}{max width=\textwidth}
\begin{tabular}{l|ccccc|ccccc|ccccc}
\hline
\multirow{2}{*}{Method}
& \multicolumn{5}{c|}{Sunny}
& \multicolumn{5}{c|}{Cloudy}
& \multicolumn{5}{c}{Sunset} \\
\cline{2-16}
& \multicolumn{1}{c}{>1} & \multicolumn{1}{c}{>3} & \multicolumn{1}{c}{>5} & \multicolumn{1}{c}{MAE} & \multicolumn{1}{c|}{RMS}
& \multicolumn{1}{c}{>1} & \multicolumn{1}{c}{>3} & \multicolumn{1}{c}{>5} & \multicolumn{1}{c}{MAE} & \multicolumn{1}{c|}{RMS}
& \multicolumn{1}{c}{>1} & \multicolumn{1}{c}{>3} & \multicolumn{1}{c}{>5} & \multicolumn{1}{c}{MAE} & \multicolumn{1}{c}{RMS} \\
\hline

\multicolumn{16}{l}{\textit{Trained on OmniThings only}} \\
\hline
OmniMVS~\cite{won2019omnimvs}
& 27.16 & 6.13 & 3.98 & 1.24 & 3.09
& 28.13 & 5.37 & 3.54 & 1.17 & 2.83
& 26.70 & 6.19 & 4.02 & 1.24 & 3.06 \\
OmniMVS+$_{32}$~\cite{won2020end}
& 13.57 & \underline{4.81} & \underline{3.10} & 0.88 & \underline{2.56}
& 13.59 & \underline{4.81} & \underline{3.15} & 0.87 & 2.53
& 13.36 & \underline{4.71} & \underline{2.93} & 0.87 & \underline{2.50} \\
S-OmniMVS~\cite{chen2023s}
& 17.19 & 6.03 & 3.89 & 1.11 & 3.60
& -- & -- & -- & -- & --
& -- & -- & -- & -- & -- \\
RomniStereo$_{64}$~\cite{deng2025omnistereo}
& \underline{11.25} & 5.30 & 3.59 & 0.80 & 2.57
& \underline{10.97} & 5.03 & 3.44 & \underline{0.73} & \underline{2.47}
& \underline{10.94} & 4.99 & 3.29 & \underline{0.72} & 2.56 \\
MDP-Omni~\cite{son2025mdp}
& 12.72 & 4.97 & 3.26 & \underline{0.79} & 2.62
& -- & -- & -- & -- & --
& -- & -- & -- & -- & -- \\
Ours
& \textbf{8.45} & \textbf{3.35} & \textbf{2.04} & \textbf{0.54} & \textbf{2.07}
& \textbf{8.69} & \textbf{3.51} & \textbf{2.25} & \textbf{0.57} & \textbf{2.15}
& \textbf{8.27} & \textbf{3.13} & \textbf{1.94} & \textbf{0.53} & \textbf{2.04} \\
\hline

\multicolumn{16}{l}{\textit{Fine-tuned on OmniHouse and Sunny}} \\
\hline
OmniMVS-ft~\cite{won2019omnimvs}
& 13.93 & 2.87 & 1.71 & 0.79 & 2.12
& 12.20 & 2.48 & 1.46 & 0.72 & 1.85
& 14.14 & 2.88 & 1.71 & 0.79 & 2.04 \\
OmniMVS+$_{32}$-ft~\cite{won2020end}
& 7.48 & 3.57 & 2.42 & 0.57 & 2.42
& 7.29 & 3.38 & 2.30 & 0.54 & 2.31
& 7.82 & 3.60 & 2.42 & 0.58 & 2.36 \\
S-OmniMVS-ft~\cite{chen2023s}
& 6.66 & 2.18 & 1.40 & 0.47 & 1.98
& -- & -- & -- & -- & --
& -- & -- & -- & -- & -- \\
RomniStereo$_{64}$-ft~\cite{deng2025omnistereo}
& 4.61 & 1.78 & 1.10 & 0.32 & 1.43
& 4.94 & 1.83 & 1.16 & 0.34 & 1.53
& \underline{4.88} & 1.90 & 1.19 & \underline{0.34} & 1.49 \\
MDP-Omni-ft~\cite{son2025mdp}
& 4.51 & \underline{1.43} & \textbf{0.86} & 0.33 & 1.43
& 5.34 & 1.68 & 1.03 & 0.36 & 1.53
& 4.90 & \underline{1.51} & \textbf{0.90} & 0.35 & 1.46 \\
Ours-ft
& \textbf{3.75} & \textbf{1.39} & \textbf{0.86} & \underline{0.28} & \underline{1.32}
& \textbf{3.93} & \textbf{1.37} & \textbf{0.84} & \textbf{0.28} & \textbf{1.29}
& \textbf{3.97} & \textbf{1.46} & \underline{0.92} & \textbf{0.29} & \underline{1.38} \\
Ours-ft (distilled)
& \underline{3.85} & \underline{1.43} & \underline{0.87} & \textbf{0.27} & \textbf{1.30}
& \underline{4.38} & \underline{1.50} & \underline{0.92} & \underline{0.29} & \underline{1.35}
& \underline{4.12} & \underline{1.52} & 0.95 & \textbf{0.29} & \textbf{1.36} \\
\hline
\end{tabular}
\end{adjustbox}
\label{tab:quantitative_sunny_cloudy_sunset}
\end{table*}

\cref{tab:quantitative_omnithings_house,tab:quantitative_sunny_cloudy_sunset} report quantitative results on OmniThings, OmniHouse, and the weather-variant datasets (Sunny, Cloudy, Sunset) using standard error metrics ($>1, >3, >5$, MAE, RMS) together with inference time. We report comparisons for (i) models trained only on OmniThings, and (ii) models further fine-tuned on OmniHouse and Sunny, following \cref{Distillation}.

\paragraph{Trained on OmniThings only.}
Our pre-trained model achieves the best overall accuracy across all three datasets.
Notably, the improvements are consistent across the thresholded error metrics, indicating that our gains are not limited to a single measure but reflect broadly improved depth reliability.
We observe similar trends on OmniHouse and the weather-variant datasets (Sunny, Cloudy, Sunset), where Ours achieves the lowest MAE/RMS among all compared methods, suggesting strong cross-domain generalization from OmniThings-only training.

\paragraph{Fine-tuned on OmniHouse and Sunny.}
Fine-tuning further strengthens the performance of Ours-ft, especially on the target domains.
Compared to the previous state-of-the-art model \cite{son2025mdp}, Ours-ft improves OmniThings MAE from $1.96$ to $1.70$ and RMS from $5.08$ to $4.69$ , demonstrating that our method retains strong performance even after adaptation.
On OmniHouse, Ours-ft achieves MAE $0.28$ and RMS $0.89$.
On Sunny, Ours-ft reduces MAE from $0.33$ to $0.28$ and RMS from $1.43$ to $1.32$, indicating robust performance under outdoor illumination changes.

\paragraph{Efficiency and distilled variant.}
We measured the inference latency of all models under a consistent environment on an NVIDIA RTX 4070 SUPER GPU. Since the source code for recent state-of-the-art model \cite{son2025mdp} is not publicly available, we derived its inference latency proportionally, based on the performance ratios reported relative to other established models. Under this setup, Ours-ft achieves superior accuracy while running in $148$\,ms.
Moreover, our distilled variant, Ours-ft (distilled), reduces inference time to $102$\,ms (about $31.1\%$ faster than Ours-ft) with only minor degradation on OmniThings and competitive accuracy on OmniHouse and Sunny.
This trade-off suggests that our approach is amenable to practical deployment under constrained compute budgets. 

\begin{figure}[!t]
    \centering
    \includegraphics[width=1.0\linewidth]{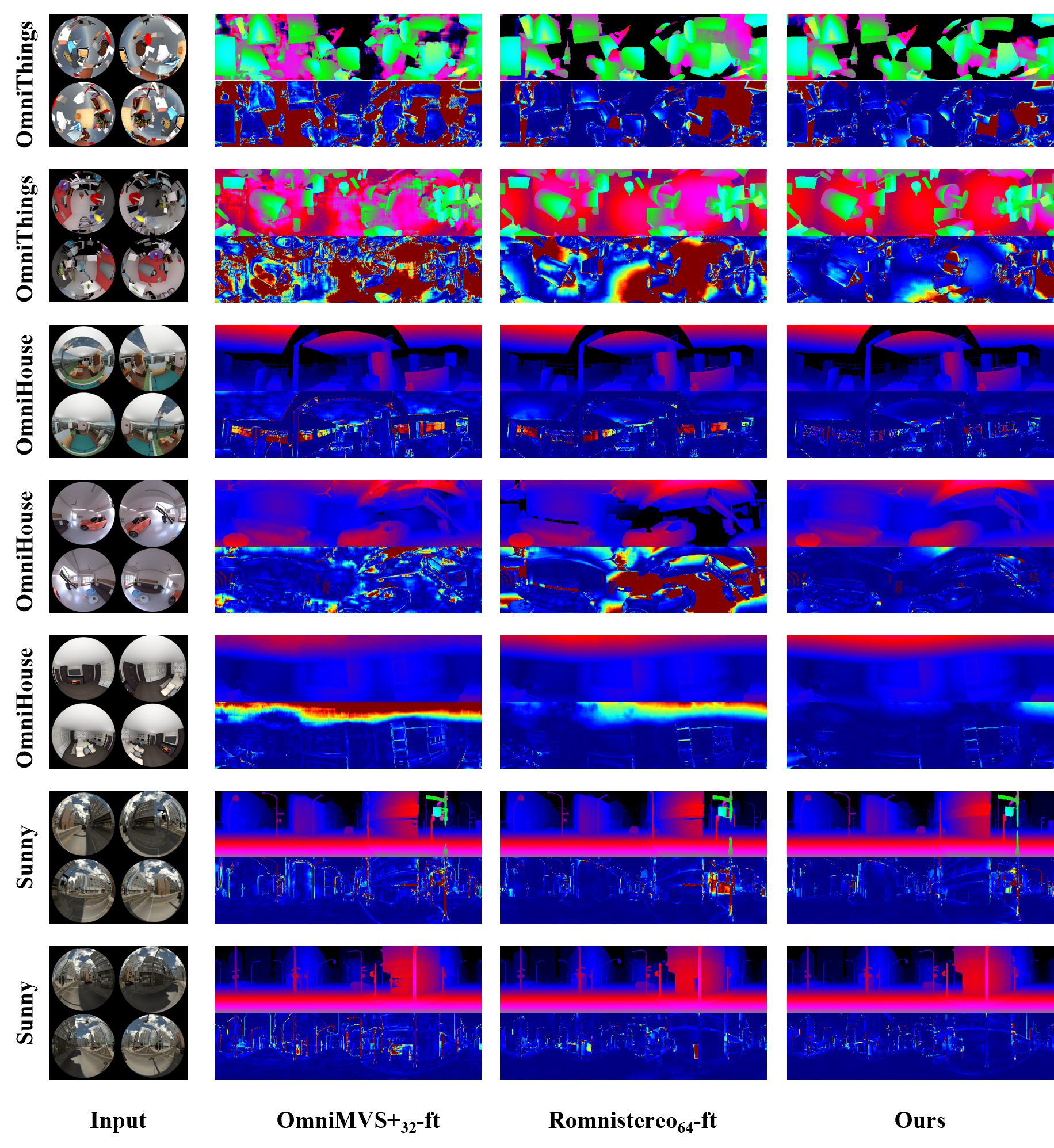}
    \caption{
    Qualitative comparison on OmniThings, OmniHouse, and Sunny. Three examples per dataset are shown. We report the strongest fine-tuned baseline variants \cite{won2019omnimvs, jiang2024RomniStereo}. The leftmost column shows the four input images; for each method, we visualize the predicted depth (top) and the error map (bottom).}
    \label{fig:qualitative}
\end{figure}

\subsection{Qualitative Results}
\label{QualitativeResults}
Fig.~\ref{fig:qualitative} compares our method with OmniMVS$_{32}$-ft and RomniStereo$_{64}$-ft on OmniThings, OmniHouse, and Sunny.
On \textbf{OmniThings}, our method produces cleaner depth around crowded object boundaries, reducing boundary artifacts visible in the baselines.
On \textbf{OmniHouse}, it better preserves large planar regions (e.g., walls/ceilings) with fewer global mispredictions.
On \textbf{Sunny}, it more reliably recovers thin structures (e.g., poles/signs) and challenging reflective/ transparent areas (e.g., windows), as reflected in the error maps.
Overall, our predictions show consistently lower errors and sharper geometry across all datasets.

\subsection{Ablation Studies}
We evaluate the contribution of each component through ablation studies on the Sunny dataset. All variants are trained for 30 epochs under an identical training protocol and evaluated using the same inverse-index metrics as in the main experiments.

\begin{table}[H]
\centering
\small
\setlength{\tabcolsep}{5pt}
\renewcommand{\arraystretch}{1.05}
\caption{Ablation on configuration choices.}
\begin{tabular}{c|ccc|ccc|ccccc}
\hline
\multirow{2}{*}{ID}
& \multicolumn{3}{c|}{Similarity}
& \multicolumn{3}{c|}{Context}
& \multicolumn{5}{c}{Metrics ($\downarrow$)} \\
\cline{2-4}\cline{5-7}\cline{8-12}
& CNN & DINO & MVC
& CNN & DINO & Def.
& $>1$ & $>3$ & $>5$ & MAE & RMS \\
\hline
(a) &       & \cmark &       &       & \cmark &       &  9.70   &  3.01  &  1.69   &  0.51  &  1.76 \\
(b) & \cmark &       &       & \cmark &       &       &  5.06   &  1.84   &  1.13   &  0.38   &  1.69 \\
(c) & \cmark &       &       &       & \cmark &       &  5.37   &  1.98   &  1.23   &  0.38   &  1.62   \\
(d) & \cmark &       &       & \cmark & \cmark &       &  4.69   &  1.69   &  \textbf{1.07}   &  0.35   &  \textbf{1.55}\\
(e) & \cmark &       & \cmark & \cmark & \cmark &       & 4.60 & 1.74 & 1.09 & 0.35 & 1.63 \\
(f) & \cmark &       & \cmark & \cmark & \cmark & \cmark & \textbf{4.49} & \textbf{1.68} & \textbf{1.07} & \textbf{0.34} & 1.59 \\
\hline
\end{tabular}
\label{tab:ablation}
\vspace{-1.5em}
\end{table}

\begin{table}[t]
\centering
\small
\setlength{\tabcolsep}{5pt}
\renewcommand{\arraystretch}{1.05}
\caption{Ablation studies for multi-view consensus modules.}
\begin{tabular}{c|ccc|ccccc}
\hline
\multicolumn{1}{c|}{ID}
& \multicolumn{3}{c|}{Modules}
& \multicolumn{5}{c}{Metrics ($\downarrow$)} \\
\cline{2-4}\cline{5-9}
& Corr & 2D Aggr. & MVC
& \multicolumn{1}{c}{>1} & \multicolumn{1}{c}{>3} & \multicolumn{1}{c}{>5}
& \multicolumn{1}{c}{MAE} & \multicolumn{1}{c}{RMS} \\
\hline
(a) & \cmark &       &          & 5.00 & 1.86 & 1.19 & 0.37 & 1.59 \\
(b) & \cmark & \cmark &         & 4.67 & 1.68 & 1.06 & 0.34 & 1.52 \\
(c) & \cmark & \cmark & \cmark  & 4.49 & 1.68 & 1.07 & 0.34 & 1.59 \\
\hline
\end{tabular}
\label{tab:ablation_gev}
\end{table}

\paragraph{Ablation on each module.}
Tab.~\ref{tab:ablation} studies how to construct and integrate the similarity volume and context branches. Among the single-context configurations (a, b, c), no single context feature is consistently best, and using both CNN and DINO as context features outperforms all of them, leading to lower threshold errors and MAE (d). Adding the MVC-based similarity volume slightly improves the >1 metric but does not consistently benefit other metrics and increases RMS (e). Finally, incorporating the default context together with CNN+DINO yields the best overall trade-off, achieving the lowest >1 and MAE while remaining competitive in RMS (f). Qualitative visualizations of the resulting ERP context—highlighting how deformable attention and different backbone combinations affect context consistency—are included in the supplementary material.
\paragraph{Ablation on similarity volumes.}
We further analyze the role of the similarity volume design in Tab.~\ref{tab:ablation_gev}.
Corr corresponds to the correlation profile described in \cref{similarityvolumes}, while MVC denotes our proposed Multi-view Consensus Volume.
Compared to the Corr baseline, introducing the full similarity volumes (2D Aggr. + MVC) improves overall performance.
Notably, MVC is more effective for fine-grained accuracy, achieving the most pronounced improvement on the tightest error threshold ($>1$) in (c), highlighting the benefit of explicitly enforcing cross-view agreement for precise similarity estimation.

\section{Conclusion}
We presented \textbf{OmniDS}, an iterative omnidirectional depth framework that replaces rigid fixed-projection aggregation in multi-fisheye systems by combining dynamic context fusion with consensus-aware multi-view similarity. A dual-stream encoder pairs CNN geometric cues with frozen DINOv3 semantic priors and reprojects them into ERP space at each refinement step, producing more coherent context around occlusions and thin structures. In parallel, a multi-view consensus volume models cross-camera agreement beyond pairwise correlation, providing stable cues for iterative refinement. A distilled MobileNet-based variant reduces inference time by 31.1\% with minimal accuracy loss, and experiments on five benchmarks show state-of-the-art performance and strong generalization across indoor, outdoor, and varying weather conditions.

A limitation is that we evaluate only on synthetic data with a fixed four-camera rig; extending our framework to arbitrary camera configurations with an unconstrained number of cameras, while accounting for calibration noise, is left for future work. Additionally, qualitative examples of failure cases are provided in the Supplementary Material.

\section{Acknowledgment}
This work was partly supported by Institute of Information \& communications Technology Planning \& Evaluation (IITP) grant funded by the Korea government (MSIT) [NO.RS-2021-II211343, Artificial Intelligence Graduate School Program (Seoul National University)] and National Research Foundation of Korea (NRF) grant funded by the Korea government (MSIT) [No. RS-2024-00359718].

\bibliographystyle{splncs04}
\bibliography{main}
\end{document}